\documentclass{article}
\usepackage{iclr2027_conference,times}
\usepackage[T1]{fontenc}
\usepackage[utf8]{inputenc}
\usepackage{graphicx}
\usepackage{amsmath,amssymb}
\usepackage{booktabs,multirow}
\usepackage[table]{xcolor}
\usepackage{adjustbox}
\usepackage{placeins}
\usepackage{float}
\usepackage{tabularx}
\usepackage{hyperref}
\usepackage{url}

\definecolor{pixvlrow}{HTML}{D9EBE3}
\definecolor{baselinerow}{HTML}{EDF0F5}
\definecolor{sectionblue}{HTML}{DBE8F6}
\definecolor{sectionlavender}{HTML}{E8E0F1}
\definecolor{maintask}{HTML}{E1EAF2}
\definecolor{mainours}{HTML}{F5F5F5}
\definecolor{linkblue}{HTML}{355C85}
\hypersetup{colorlinks=true,allcolors=linkblue,pdfauthor={Yicheng Xiao, Haoxuan Ma, Caorui Li, Yucheng Wu, Weijie Wang, Haoxiao Wang, Shuang Chen, Fan Yang, Haiyun Guo, Jinqiao Wang},pdftitle={PixVL: Self-Supervised Training of Pixel-Level MLLMs via a Unified Mask--Text Consistency Cycle}}

\iclrfinalcopy
\title{\protect\resizebox{\textwidth}{!}{PixVL: Self-Supervised Training of Pixel-Level MLLMs via a Unified Mask--Text}\\
Consistency Cycle}
\author{Yicheng Xiao\textsuperscript{1,*}, Haoxuan Ma\textsuperscript{2,*},
Caorui Li\textsuperscript{3}, Yucheng Wu\textsuperscript{4}, Weijie Wang\textsuperscript{5},\\
Haoxiao Wang\textsuperscript{5}, Shuang Chen\textsuperscript{6},
Fan Yang\textsuperscript{1}, Haiyun Guo\textsuperscript{1,\ensuremath{\dagger}},
Jinqiao Wang\textsuperscript{1}}
\input{arxiv_title}

\begin{document}
\raggedbottom
\setlength{\parskip}{4.5pt}
\setcounter{topnumber}{1}
\maketitle
\begin{abstract}
Unified pixel-level multimodal large language models (MLLMs) support both region understanding and region segmentation, but their joint training is limited by scarce mask--text pairs and interference between the two objectives. We introduce PixVL, a self-supervised post-training framework that turns Mask-to-Text (M2T) and Text-to-Mask (T2M) into mutual generators and verifiers on mask-only data. A direct mask--text--mask cycle is insufficient: reconstruction IoU can change abruptly when a discriminative attribute is omitted and can reward underspecified descriptions. PixVL therefore combines geometric reconstruction with confuser-aware semantic verification. A generated caption is evaluated by whether it identifies the target among similar candidate regions, receiving the model's target confidence only when the choice is correct. Verification across temporally separated video views further discourages position and shape shortcuts. To couple the two learning directions, the highest-reward caption conditions a new group of T2M rollouts, and its verified reward scales the T2M advantages after group normalization. Segmentation credit is restricted to mask-token positions, and both objectives update a shared policy in one optimizer step. Starting from SAM-Tok-4B, PixVL uses approximately 20k choose-one cold-start examples and 250k mask-only cycle-training samples. It improves GroundingSuite gIoU from 56.1 to 64.8 and the reported DLC-Bench average from 65.6 to 69.7, while also improving referring-expression segmentation and grounded conversation generation. These results support self-verified bidirectional learning as a way to scale pixel-level vision--language alignment beyond paired regional descriptions.

\end{abstract}
\section{Introduction}
\label{sec:introduction}

Multimodal large language models (MLLMs)~\citep{deepmind2025gemini25pro,bai2025qwen3vl,wang2025internvl35,hong2025glm45v,kimi2025kimivl} increasingly understand images, but precise interaction requires identifying particular objects, parts, and regions. Masks provide a shared interface: Mask-to-Text (M2T) describes a selected region, while Text-to-Mask (T2M) localizes a region specified in language. Segmentation models such as LISA~\citep{lai2024lisa} and VISA~\citep{yan2024visa} emphasize T2M; regional understanding models such as DAM~\citep{lian2025describe}, PixelRefer~\citep{yuan2025pixelrefer}, and GAR~\citep{wang2026gar} emphasize M2T. Unified models, including Sa2VA~\citep{yuan2025sa2va} and SAM-Tok~\citep{zhou2026samtok}, support both directions.

\begin{figure}[t]
\centering
\includegraphics[width=\linewidth]{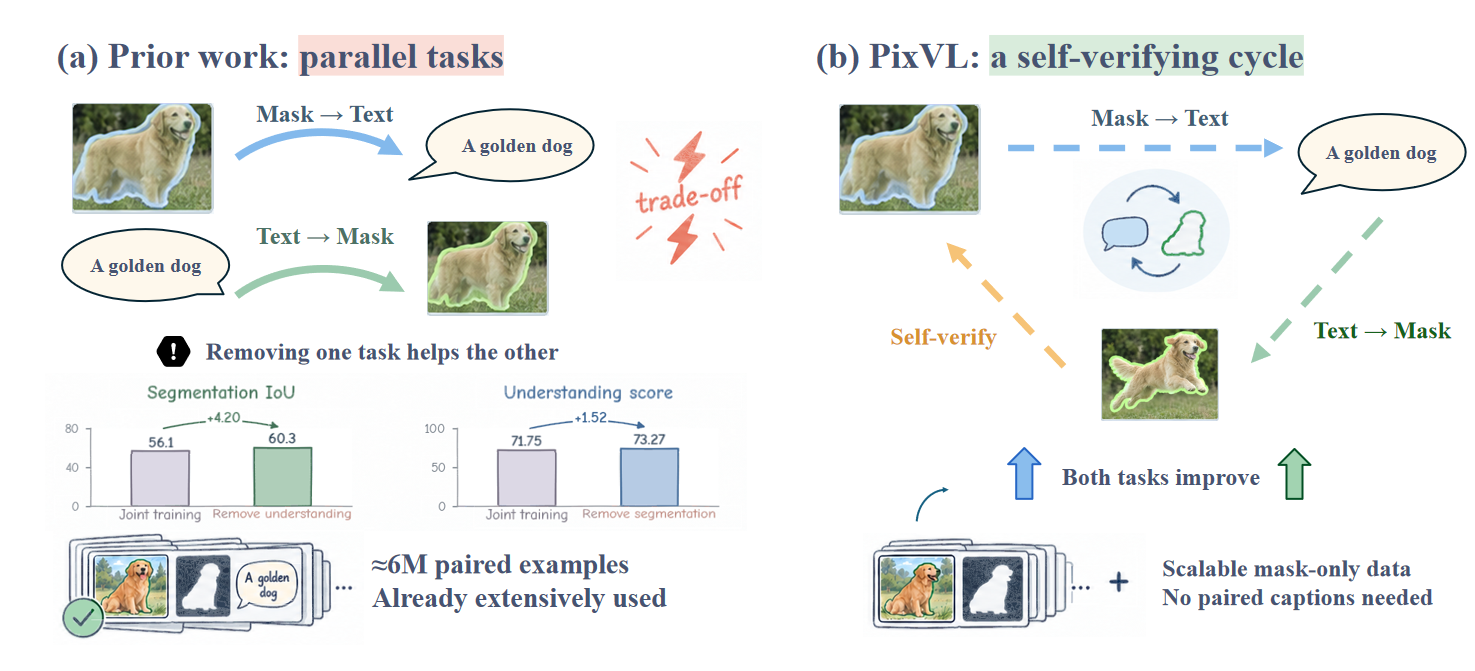}
\caption{Motivation and central idea of PixVL. (a) Prior pixel-level MLLMs train Mask-to-Text and Text-to-Mask as parallel tasks. Our SAM-Tok ablation reveals a trade-off: removing either training task improves the other, as shown by the two bar charts using their respective metrics. The limited pool of approximately 6M paired examples has already been extensively used. (b) PixVL connects the complementary directions in a self-verifying cycle, using scalable mask-only data to improve both tasks without additional paired captions in the cycle-training stage. The illustrated different views represent video samples; static samples reuse the same image and mask.}
\label{fig:motivation}
\end{figure}

However, architectural unification does not ensure effective joint learning. Existing models train the directions as parallel tasks, although language-sequence and mask-specific supervision differ in form and density. Our SAM-Tok ablation reveals a \emph{training trade-off}: removing understanding training raises segmentation IoU from 56.1 to 60.3, whereas removing segmentation training raises the reported understanding score from 71.75 to 73.27 (Figure~\ref{fig:motivation}(a)). The roughly six million available mask--text examples have already been extensively used, and annotating the attributes and relations that distinguish similar instances is costly. Image and video datasets instead offer abundant masks without paired language. Can these resources make the two directions reinforce each other?

We first tried a simple mask--text--mask cycle. Given an image and target mask, M2T generates a caption; T2M uses that caption to reconstruct the mask in the same image; reconstruction IoU then rewards the caption. A good description should recover its source region, providing supervision without a human-written caption. \emph{Unexpectedly, this direct cyclic training degraded performance, especially referring-expression segmentation.} The geometric-reward-only ablation in Table~\ref{tab:ablation} confirms that reconstruction rewards alone are insufficient.

\begin{figure}[t]
\centering
\includegraphics[width=\linewidth]{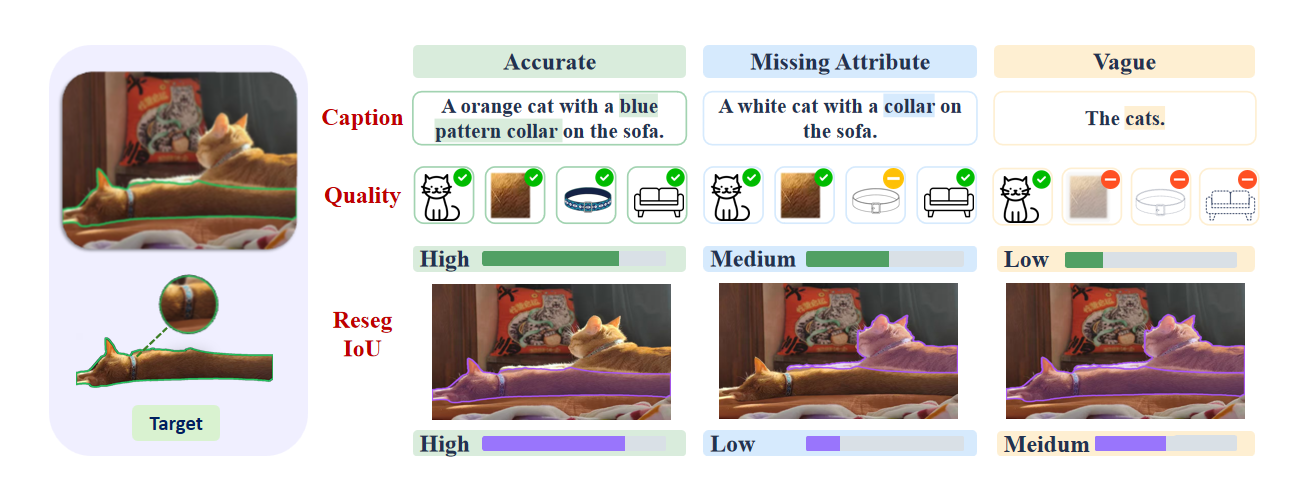}
\caption{Caption quality and re-segmentation IoU can have different rankings. The target is the orange cat with a blue patterned collar (left, green outline). Purple overlays illustrate recovering the target, the other cat, or both cats. Caption quality decreases from accurate to missing-attribute to vague, whereas IoU is high, low, and medium, respectively. Icons summarize descriptive details; bars indicate qualitative levels, not measured scores. This mismatch motivates semantic verification beyond geometric overlap.}
\label{fig:iou-failure}
\end{figure}

Investigating this failure led us to the central mismatch: \emph{re-segmentation IoU does not reliably track caption quality}. The mask--language mapping is many-to-many. In Figure~\ref{fig:iou-failure}, the accurate caption identifies the orange cat by its blue patterned collar. The middle caption changes the color to white and omits the collar pattern; selecting the other cat yields low IoU. The vague ``The cats.'' loses more target-specific information, yet selecting both cats produces moderate overlap. Thus, caption quality decreases from high to medium to low, while IoU changes from high to low to medium. IoU remains useful for checking reconstruction, but its ranking does not establish descriptive sufficiency.

To address this mismatch, we introduce \textbf{PixVL}, which adds \emph{confuser-aware semantic verification} to the cycle. Instead of checking only the recovered shape, the model must identify the target among similar candidate regions. We combine same-image structural confusers with similar regions retrieved from a mask embedding bank. The shared model chooses the region matching each caption, earning its target confidence only when the choice is correct. Together with mask and box IoU, this rewards descriptions that distinguish the target and recover its geometry.

A second problem remains: even a discriminative same-image caption can exploit \emph{position or shape shortcuts}. A phrase such as ``the upper-left region'' may identify a mask without learning persistent attributes. We therefore verify captions across temporally separated observations of the same video track: M2T describes the first view, and re-segmentation, re-grounding, and choose-one verification operate in the second. Changes in position, scale, pose, and context make image-specific shortcuts less reliable. A small proportion of video pairs supplies this constraint, while static samples reuse their image and mask to retain broad mask-only coverage.

Finally, PixVL turns the best verified caption into reverse supervision. Its quality controls T2M update strength through post-normalization advantage scaling, and segmentation credit is restricted to mask-token positions. Both directions contribute to one shared policy update. The completed cycle improves segmentation and understanding together using 250k mask-only training samples.

\begin{samepage}
Our contributions are threefold:
\begin{itemize}
\item We expose the failure of direct geometric cycle training and introduce a self-supervised mask--text consistency cycle that improves both segmentation and understanding beyond extensively used paired data.
\item We combine confuser-aware semantic verification with cross-view reconstruction to address the caption--IoU mismatch and image-specific shortcuts.
\item We develop best-caption reverse supervision with post-normalization quality coupling and mask-token credit, connecting the two directions in one shared policy update.
\end{itemize}
\end{samepage}

\section{Related Work}
\label{sec:related-work}

\subsection{Pixel-Level Multimodal Large Language Models}

Existing pixel-level MLLMs can be grouped by the role of masks in the model:
Region Segmentation, Region Understanding, and unified bidirectional models. Region Segmentation
methods such as LISA~\citep{lai2024lisa} and VISA~\citep{yan2024visa} combine
language reasoning with segmentation modules to support referring and
reasoning segmentation. They primarily generate masks from language using
manually constructed text--region supervision, without exploiting the reverse
region-understanding direction as a source of training signals. VideoSeg-R1~\citep{xu2026videosegr1} additionally uses reinforcement learning for reasoning video segmentation.

Region Understanding methods such as DAM~\citep{lian2025describe},
PixelRefer~\citep{yuan2025pixelrefer}, and GAR~\citep{wang2026gar} instead treat
a mask as an input prompt for captioning, question answering, or fine-grained
attribute analysis. Regional understanding must distinguish the target from
similar objects using both local evidence and context, and thus depends heavily
on scarce, linguistically diverse mask--text pairs. PixCLIP~\citep{xiao2026pixclip} further studies fine-grained pixel--text alignment across granularities.

Recent work such as Sa2VA~\citep{yuan2025sa2va} and
SAM-Tok~\citep{zhou2026samtok} further unifies mask input and mask
output in one model. While these interfaces enable bidirectional learning,
prior work emphasizes architectural unification and multi-task supervision
rather than using the two capabilities as mutual data generators and verifiers.
PixVL instead builds a scalable self-supervised loop on mask-only data.

\subsection{Self-Training and Cycle Consistency}

Self-training and pseudo-labeling use a model's own predictions as supervision
across semi-supervised vision and vision--language learning
\citep{xie2020noisy}. Cycle consistency uses forward and reverse mappings to
recover the input when paired annotations are unavailable
\citep{zhu2017cyclegan}. For Mask-to-Text and Text-to-Mask, a caption generated
from a mask can reconstruct that mask, creating pseudo mask--text pairs.

The mapping between pixels and language, however, is inherently many-to-many.
Directly minimizing a geometric reconstruction error conflates whether the
cycle returns to the original mask with whether the description is correct and
discriminative. PixVL therefore does not assume strict invertibility:
correctness-gated choose-one confidence measures semantic sufficiency, while re-segmentation and
re-grounding test geometric validity in another view. Claim-level video-caption rewards~\citep{gao2026cure} likewise use semantic verification beyond holistic response scoring.

Video mask tracks and image transformations are also widely used for temporal
consistency and cross-view self-supervision
\citep{wang2019cycle,chen2020simclr}. PixVL uses video mask tracks to reduce position and shape shortcuts: descriptions
are verified across changes in background, scale, pose, and position. Static
image samples reuse their original observation. Beyond masks, OmniReasoner~\citep{chen2026omnireasoner} learns selective inspection of long audio--video streams, while OmniVideoBench~\citep{li2026omnivideobench} evaluates complementary audio--visual reasoning.

\section{Problem Formulation}
\label{sec:problem-formulation}

Let $I$ be an image, $M$ its target mask, and $c$ a regional caption. A unified policy $\pi_\theta$, parameterized by $\theta$, supports two conditional generation interfaces:
\begin{equation}
 c\sim\pi_\theta^{\mathrm{M2T}}(\cdot\mid I,M),
 \qquad \widehat M\sim\pi_\theta^{\mathrm{T2M}}(\cdot\mid I,c).
 \label{eq:bidirectional-model}
\end{equation}
The superscripts indicate the task prompt, not separate model parameters; $\widehat M$ denotes the decoded predicted mask. Training uses images with masks or video mask tracks, without paired captions in the cycle stage. We seek descriptions that identify their source regions and provide reliable conditions for reverse segmentation. ``Self-supervised'' refers to this large-scale cycle: the model still requires mask annotations and a small supervised choose-one initialization.

\section{Method}
\label{sec:method}

\begin{figure}[!t]
\centering
\includegraphics[width=0.97\textwidth]{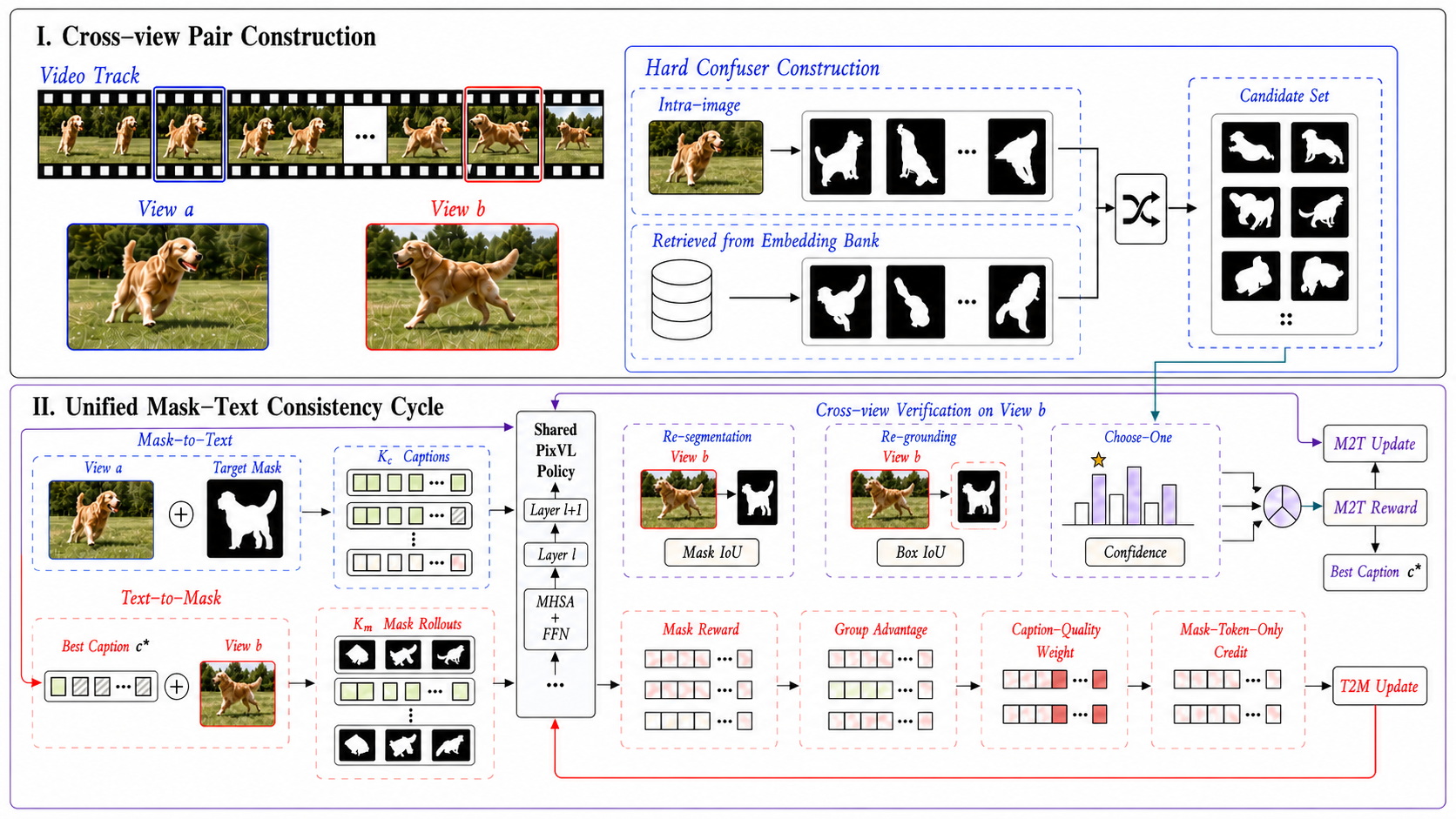}
\caption{Overview of the PixVL Framework. Temporally
separated views and hard confusers verify M2T captions using mask IoU, box IoU,
and correctness-gated choose-one confidence. The best caption conditions T2M
rollouts, and its reward weights the mask-token advantages. Both objectives
update the shared policy in one optimizer step.}
\label{fig:overview}
\end{figure}

PixVL learns through a verified mask--text cycle (Figure~\ref{fig:overview}). Captions generated from a source region are checked against another observation and hard alternative regions; the best verified caption then supervises reverse segmentation. All generation and verification in an iteration use one pre-update policy snapshot, denoted by $\pi_{\theta_{\mathrm{old}}}$. Both directions are collected before a single shared update, so the reverse group is conditioned on the caption selected by the same policy.

\subsection{Cross-View Verification Context}
\label{sec:cross-view}

Let $(I^a,M^a)$ and $(I^b,M^b)$ denote the source and verification observations of one region. For a video track, they are $(I_t,M_t)$ and $(I_{t+\Delta},M_{t+\Delta})$, where $t$ is the source frame and $\Delta$ the temporal interval. M2T observes only view $a$; caption verification and reverse segmentation use view $b$. Motion and viewpoint changes make descriptions based only on coordinates or approximate shape less reliable, requiring information that persists with the tracked instance.

We use SA-V~\citep{ravi2025sam2} tracks with $\Delta=8$; video pairs comprise 20\% of cycle samples. Static SA-1B~\citep{kirillov2023segment} samples set $(I^b,M^b)=(I^a,M^a)$. This mixture combines temporal constraints with broad image coverage, without requiring synthetic transformations.

\subsection{Hard Confusers and Verifier Initialization}
\label{sec:confuser-verifier}

\begin{figure}[t]
\centering
\includegraphics[width=\linewidth]{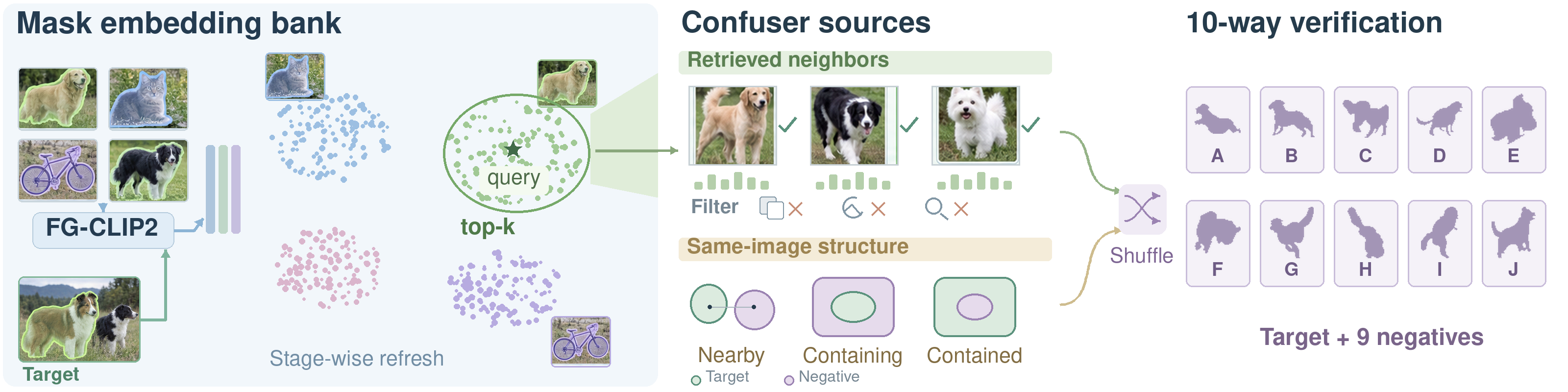}
\caption{Hard-confuser construction. FG-CLIP2 indexes masked regions and queries a stage-wise refreshed bank. Filtered top-$k$ neighbors and same-image structural negatives form nine confusers; adding the target gives a shuffled ten-way input. The dot cloud, structural set diagrams, and mask-option glyphs are schematic. Green and purple distinguish the target and negative only in the structural diagrams; all verifier options use the same styling.}
\label{fig:confusers}
\end{figure}

For view $b$, we construct a candidate set $C^b$ containing the target and nine confusers (Figure~\ref{fig:confusers}). Same-image negatives include nearby masks, masks containing the target, and masks contained within it. Nearby masks test instance identity; containing and contained masks test whether a caption refers to the whole region, its surroundings, or a part. This makes correct spatial granularity necessary for target discrimination.

For cross-image negatives, FG-CLIP2~\citep{xie2025fgclip2} embeds masked regions into a stage-wise refreshed bank. We encode the target with the same encoder, retrieve its top-$k$ neighbors, and filter duplicates, invalid masks, and irrelevant regions; $k$ denotes the retrieval-pool size. Retained neighbors are mixed with same-image negatives to obtain nine confusers. Each candidate keeps its own visual context. Cross-image retrieval supplies visually similar alternatives that need not occur in the target scene. These alternatives test fine-grained appearance beyond image layout, complementing the instance and granularity distinctions imposed by nearby, containing, and contained masks. We prefer nearby valid masks for same-image sampling, discard samples with insufficient candidates, and shuffle all ten options to remove fixed answer-position cues.

The shared model receives approximately 20k choose-one SFT examples, each comprising an image, candidate masks, a caption, and its matching option. Randomized order, sources, and difficulty discourage mask-area and option-frequency shortcuts. Before this initialization, the model may generate descriptions without reliably comparing candidate masks or following the option-answer format. This limited SFT establishes candidate comparison and stable option probabilities, without introducing a separate reward model or providing paired captions for the 250k cycle samples. During RL, verification uses the current policy without gradients, and its parameters advance with the shared update.

\subsection{Confuser-Aware Caption Verification}
\label{sec:caption-verification}

For each source region, we sample a group of $K_c$ captions and obtain one independent verification mask per caption:
\begin{equation}
\begin{aligned}
 c_k &\sim \pi_{\theta_{\mathrm{old}}}^{\mathrm{M2T}}(\cdot\mid I^a,M^a),\qquad k=1,\ldots,K_c,\\
 \widetilde M_k^b &\sim \pi_{\theta_{\mathrm{old}}}^{\mathrm{T2M}}(\cdot\mid I^b,c_k).
\end{aligned}
\label{eq:caption-rollouts}
\end{equation}
The tilde distinguishes these verification predictions from the later reverse-training rollouts. They check reconstruction, but Figure~\ref{fig:iou-failure} shows why overlap alone cannot assess a caption's discriminative content.

The choose-one interface assigns probability $p_k(z)$ to option $z$ given $I^b$, $C^b$, and $c_k$. With $z^+$ denoting the shuffled target index, semantic verification is
\begin{equation}
 \widehat z_k=\arg\max_{z\in\{1,\ldots,10\}}p_k(z),
 \qquad r_{\mathrm{choice}}(c_k)=p_k(z^+)\,\mathbf 1[\widehat z_k=z^+].
 \label{eq:choice}
\end{equation}
Here $\widehat z_k$ is the selected option and $\mathbf 1[\cdot]$ is the indicator. Correct choices earn target confidence, distinguishing decisive from uncertain identification; incorrect choices receive no semantic reward. This check requires no unique reference sentence: different descriptions can succeed if they distinguish the annotated target. Its scope is the constructed candidate set, so geometric evidence remains necessary to check whether the caption also supports faithful reconstruction.

We combine this discriminative signal with box and mask reconstruction:
\begin{equation}
\begin{aligned}
 r_k^c={}&\alpha\,\operatorname{IoU}\!\left(B(\widetilde M_k^b),B(M^b)\right)\\
        &+\beta\,\operatorname{IoU}(\widetilde M_k^b,M^b)
         +\gamma\,r_{\mathrm{choice}}(c_k).
\end{aligned}
 \label{eq:m2t-reward}
\end{equation}
$B$ extracts a tight bounding box; the first two terms are re-grounding and re-segmentation rewards from the same prediction, requiring no extra detector. Box overlap checks location and extent with greater tolerance to boundary errors, whereas mask overlap evaluates pixel reconstruction. Neither is treated as a complete language-quality score. The weights are $(\alpha,\beta,\gamma)=(0.25,0.25,0.5)$, and superscript $c$ denotes the caption direction. An incorrect choice removes only the semantic term. Malformed responses, failed decoding, and empty masks receive zero total reward; caption groups with insufficient quality or negligible reward variation are skipped.

\subsection{Quality-Coupled Reverse Supervision}
\label{sec:quality-coupling}

A retained caption group supplies both M2T feedback and a common language condition for T2M. We select
\begin{equation}
 k^\star=\arg\max_{1\le k\le K_c}r_k^c,
 \qquad c^\star=c_{k^\star},\qquad w=r_{k^\star}^c,
 \label{eq:best-caption}
\end{equation}
where $k^\star$ is the selected index and $w\in[0,1]$ its verified quality. With $(I^b,c^\star)$ fixed, the same policy generates $K_m$ fresh mask predictions:
\begin{equation}
 \widehat M_j^b\sim\pi_{\theta_{\mathrm{old}}}^{\mathrm{T2M}}(\cdot\mid I^b,c^\star),
 \qquad r_j^m=\operatorname{IoU}(\widehat M_j^b,M^b),\quad j=1,\ldots,K_m.
 \label{eq:t2m-rollouts}
\end{equation}
Here $K_m$ is the reverse-group size, $\widehat M_j^b$ the decoded prediction, and superscript $m$ the mask direction. The earlier verification masks cannot form this group because they were conditioned on different captions. No intermediate update or additional best-caption threshold is applied. The selected description is reused as input, but its verification prediction is not reused as the reverse rollout group. This separation preserves comparisons among mask completions under one fixed language condition.

Let $A_k^c$ and $A_j^m$ be the group-relative advantages obtained by independently standardizing each direction's rewards, with numerical stabilization, as in GRPO~\citep{shao2024deepseekmath}. Even a weak caption can produce relative winners within its mask group; independent normalization alone therefore does not control that group's absolute reliability. PixVL couples the directions through
\begin{equation}
 \widetilde A_k^c=A_k^c,
 \qquad \widetilde A_j^m=r_{k^\star}^c A_j^m=w A_j^m.
 \label{eq:reward-coupling}
\end{equation}
The tildes denote the advantages after coupling. Applying $w$ \emph{after} normalization attenuates unreliable reverse supervision while preserving within-group signs and rankings; scaling raw rewards beforehand would largely cancel. The coupling is asymmetric: reverse rewards do not recompute caption advantages. Rewards, selection, and advantages remain fixed during optimization, with no gradient through verification. Groups with negligible reward variation contribute no update.

\subsection{Mask-Token Credit and Joint Policy Optimization}
\label{sec:policy-objective}

Caption quality supervises the entire description; segmentation quality should directly supervise the tokens encoding the mask. For direction $d\in\{c,m\}$, let $y_i^d$ be completion $i$ under prompt $q^c=(I^a,M^a)$ or $q^m=(I^b,c^\star)$. Its token and preceding prefix at position $t$ are $y_{i,t}^d$ and $y_{i,<t}^d$. Credit is supported on
\begin{equation}
 S_i^c=\{1,\ldots,|y_i^c|\},
 \qquad S_i^m=\{t:y_{i,t}^m\in\mathcal V_{\mathrm{mask}}\},
 \label{eq:token-support}
\end{equation}
where $\mathcal V_{\mathrm{mask}}$ contains mask-start, mask-code, and mask-end tokens, and $|y|$ denotes completion length. Other T2M tokens remain in the autoregressive context but receive no direct segmentation credit. For example, JSON delimiters and ordinary explanatory words are excluded from direct T2M credit, even when they precede a mask token. This changes the loss support, not the full causal context or the set of trainable shared parameters. Completions without supported tokens are excluded; $\mathcal I_d$ denotes the valid indices.

The implemented ratio aggregates policy changes over this support:
\begin{equation}
 \rho_i^d(\theta)=\exp\!\left[
 \frac{1}{|S_i^d|}\sum_{t\in S_i^d}
 \log\frac{\pi_\theta(y_{i,t}^d\mid q^d,y_{i,<t}^d)}
               {\pi_{\theta_{\mathrm{old}}}(y_{i,t}^d\mid q^d,y_{i,<t}^d)}
 \right],
 \label{eq:mask-token-logprob}
\end{equation}
where $|S_i^d|$ counts supported positions. This is the exponentiated difference of length-normalized sequence log-probabilities, with the pre-update denominator fixed. Thus, clipping acts on a sequence ratio while the support determines which predictions receive credit.

For each direction, the unweighted clipped group-relative loss is
\begin{equation}
 \mathcal L_d(\theta)=-\frac{1}{|\mathcal I_d|}\sum_{i\in\mathcal I_d}
 \min\!\left\{\rho_i^d(\theta)A_i^d,
 \operatorname{clip}\!\left(\rho_i^d(\theta),1-\epsilon,1+\epsilon\right)A_i^d\right\},
 \label{eq:directional-objective}
\end{equation}
Here $\epsilon=0.2$ is the clipping radius: $\operatorname{clip}$ restricts the ratio to $[1-\epsilon,1+\epsilon]$. An empty valid set contributes zero. The loss averages valid completions within each group. In particular, a longer response does not receive extra weight merely by containing more tokens: its ratio is normalized over its own support before the completion-level surrogate is evaluated. Caption and mask groups remain separately averaged. Using the caption-quality coupling in Eq.~\eqref{eq:reward-coupling} yields the shared objective
\begin{equation}
 \mathcal L_{\mathrm{PixVL}}(\theta)
 =\mathcal L_c(\theta)+w\,\mathcal L_m(\theta).
 \label{eq:joint-update}
\end{equation}
Because $w\geq0$ is fixed, this is equivalent to substituting $\widetilde A_j^m$ into the mask loss; it applies quality once. We average the joint loss over samples and take one optimizer step after both rollout groups and all verification are complete. T2M evaluates generated language, while verified language controls reverse mask learning, closing the cycle within one shared policy.

\begin{table}[!t]
\caption{Main results on region segmentation and understanding. GroundingSuite reports gIoU; RefCOCO/+/g report validation cIoU. GCG reports METEOR (MET), CIDEr (CID), AP50, mIoU, and recall (Rec). Higher is better; bold marks the best reported score in each column. Dashes denote unavailable results.}
\label{tab:main-results}\label{tab:grounding-suite}\label{tab:refcoco}\label{tab:gcg}
\centering\fontsize{8.5}{10.2}\selectfont
\renewcommand{\arraystretch}{1.35}
\setlength{\parskip}{0pt}
\setlength{\tabcolsep}{2pt}
\vbox{\offinterlineskip\hbox{\begin{tabularx}{\linewidth}{>{\raggedright\arraybackslash}p{88pt}>{\centering\arraybackslash}p{30pt}*{5}{>{\centering\arraybackslash}p{22.5pt}}*{3}{>{\centering\arraybackslash}X}}
\toprule
\rowcolor{maintask}
\multicolumn{10}{l}{\rule[-4pt]{0pt}{18pt}\fontsize{10}{12}\selectfont\textbf{(a) Region Segmentation}} \\
\midrule[0.5pt]
& & \multicolumn{5}{c}{\textbf{GroundingSuite}} & RefCOCO & RefCOCO+ & RefCOCOg \\
\cmidrule[0.5pt](lr){3-7}\cmidrule[0.5pt](lr){8-8}\cmidrule[0.5pt](lr){9-9}\cmidrule[0.5pt](lr){10-10}
\textbf{Method} & \textbf{Size} & Stuff & Part & Multi & Single & All & Val & Val & Val \\
\midrule
LISA~\citeyearpar{lai2024lisa} & 7B & 85.2 & 21.2 & \textbf{71.5} & 42.8 & 57.6 & 74.1 & 62.4 & 66.4 \\
GLaMM~\citeyearpar{rasheed2024glamm} & 7B & 86.9 & 16.5 & 70.4 & 42.1 & 57.2 & 79.5 & 72.6 & 74.2 \\
InstructSeg~\citeyearpar{wei2025instructseg} & 3B & 56.2 & \textbf{24.2} & 66.8 & 51.3 & 52.5 & 81.8 & 77.4 & 79.3 \\
Sa2VA~\citeyearpar{yuan2025sa2va} & 4B & 90.7 & 18.3 & 73.7 & 50.0 & 62.1 & 82.4 & 77.6 & \textbf{79.7} \\
PaDT Pro~\citeyearpar{su2025padt} & 3B & 28.3 & 10.3 & 44.8 & 32.3 & 31.1 & 81.3 & 77.6 & 78.1 \\
Text4Seg~\citeyearpar{lan2025text4seg} & 8B & 85.6 & 10.2 & 51.7 & 25.6 & 45.4 & 79.2 & 72.8 & 74.0 \\
Qwen3-VL-Seg~\citeyearpar{yao2026qwen3vlseg} & 4B & -- & -- & -- & -- & -- & 82.3 & 76.2 & 78.2 \\
SAM-Tok~\citeyearpar{zhou2026samtok} & 4B & 80.4 & 12.6 & 57.6 & 52.2 & 56.1 & 82.2 & 77.2 & 78.3 \\
\rowcolor{mainours}
\textbf{PixVL} & 4B & \textbf{91.6} & 21.5 & 66.1 & \textbf{60.0} & \textbf{64.8} & \textbf{82.7} & \textbf{77.9} & 78.7 \\
\specialrule{0.65pt}{2pt}{0pt}
\end{tabularx}}
\hbox{\setlength{\tabcolsep}{2pt}\begin{tabularx}{\linewidth}{>{\raggedright\arraybackslash}p{88pt}>{\centering\arraybackslash}p{30pt}*{10}{>{\centering\arraybackslash}X}}
\rowcolor{maintask}
\multicolumn{12}{l}{\rule[-4pt]{0pt}{18pt}\fontsize{10}{12}\selectfont\textbf{(b) Region Understanding}} \\
\midrule[0.5pt]
& & \multicolumn{5}{c}{\textbf{GCG Val}} & \multicolumn{5}{c}{\textbf{GCG Test}} \\
\cmidrule[0.5pt](lr){3-7}\cmidrule[0.5pt](lr){8-12}
\textbf{Method} & \textbf{Size} & MET & CID & AP50 & mIoU & Rec & MET & CID & AP50 & mIoU & Rec \\
\midrule
LISA~\citeyearpar{lai2024lisa} & 7B & 13.0 & 33.9 & 25.2 & 62.0 & 36.3 & 12.9 & 32.2 & 24.8 & 61.7 & 35.5 \\
GLaMM~\citeyearpar{rasheed2024glamm} & 7B & 16.2 & 47.2 & 30.8 & 66.3 & 41.8 & 15.8 & 43.5 & 29.2 & 65.6 & 40.8 \\
OMG-LLaVA~\citeyearpar{zhang2024omgllava} & 7B & 14.9 & 41.2 & 29.9 & 65.6 & -- & 14.5 & 38.5 & 28.6 & 64.7 & -- \\
Sa2VA~\citeyearpar{yuan2025sa2va} & 8B & 16.4 & 49.5 & 33.2 & 67.7 & 45.1 & 16.2 & 49.0 & 32.2 & 66.8 & 44.5 \\
SAM-Tok~\citeyearpar{zhou2026samtok} & 4B & 16.1 & 48.2 & 34.7 & 69.4 & 46.6 & 16.4 & 51.4 & 34.4 & 68.4 & 48.3 \\
\rowcolor{mainours}
\textbf{PixVL} & 4B & \textbf{17.8} & \textbf{55.0} & \textbf{36.2} & \textbf{69.5} & \textbf{49.9} & \textbf{17.3} & \textbf{53.8} & \textbf{35.1} & \textbf{68.6} & \textbf{49.9} \\
\bottomrule
\end{tabularx}}}
\par\smallskip
\raggedright\footnotesize
Results are from the cited papers or the original benchmark comparisons; model variants can differ across tasks. LISA's added RefCOCO/+/g entries use its standard 7B row, without additional ReferSeg fine-tuning.
\end{table}

\section{Experiments}
\label{sec:experiments}

\paragraph{Experimental setup.}
We initialize from SAM-Tok-4B~\citep{zhou2026samtok}, use 20k choose-one cold-start examples, and train on 250k mask-only SA-1B/SA-V samples with an 8:2 image--video ratio. Both rollout groups have size $K_c=K_m=12$. We evaluate segmentation on GroundingSuite~\citep{hu2025groundingsuite} and RefCOCO/+/g~\citep{yu2016modeling,mao2016generation}, and understanding on DLC-Bench~\citep{lian2025describe} and grounded conversation generation (GCG)~\citep{rasheed2024glamm}. Appendix~\ref{app:implementation-details} provides the full training configuration.

\subsection{Joint Improvements in Segmentation and Understanding}

\paragraph{Region segmentation.}
PixVL improves GroundingSuite gIoU from 56.1 to 64.8, with gains across stuff, parts, multiple regions, and single regions (Table~\ref{tab:main-results}(a)). This is the highest overall score in the comparison, although specialized methods retain stronger scores in some categories. RefCOCO/+/g also improve from 82.2/77.2/78.3 to 82.7/77.9/78.7. The same mask-only cycle therefore improves broad grounding while preserving and modestly strengthening standard referring-expression segmentation. On GroundingSuite, the gains are 11.2 points for stuff and 8.9 for parts, alongside improvements on both single- and multi-region queries. These complementary categories suggest broader region coverage rather than an improvement confined to salient whole objects.

\paragraph{Region understanding.}
DLC-Bench average accuracy rises from 65.6 to 69.7, with both positive and negative subsets improving (Table~\ref{tab:dlc-bench}). On GCG validation, METEOR/CIDEr increase from 16.1/48.2 to 17.8/55.0 and AP50/recall from 34.7/46.6 to 36.2/49.9; test results follow the same trend, while mIoU changes are small (Table~\ref{tab:main-results}(b)). These concurrent gains support bidirectional learning rather than exchanging segmentation quality for language quality. The smaller GCG mask-IoU changes distinguish better referring descriptions and recovery of regions from uniformly better boundaries. Appendix~\ref{app:more-experiments} additionally reports gains on GRES and average MR-PACO performance; the latter is modest and not uniform across rounds.

\begin{table}[t]
\renewcommand{\arraystretch}{1.20}
\begin{minipage}[t]{.485\linewidth}\vspace{0pt}
\caption{DLC-Bench accuracy (\%).}
\label{tab:dlc-bench}
\centering
\footnotesize
\setlength{\tabcolsep}{2.5pt}
\begin{tabularx}{\linewidth}{@{}>{\raggedright\arraybackslash}Xlrrr@{}}
\toprule
Method & Size & Pos. & Neg. & Avg. \\
\midrule
GPT-4o~\citeyearpar{openai2024gpt4o}
           & --  & 43.4 & 79.6 & 61.5 \\
o1~\citeyearpar{openai2024o1}
           & --  & 46.3 & 78.8 & 62.5 \\
Claude 3.7 Sonnet~\citeyearpar{anthropic2025claude37}
           & --  & 21.8 & 50.4 & 36.1 \\
Gemini 2.5 Pro~\citeyearpar{deepmind2025gemini25pro}
           & --  & 36.5 & 75.2 & 55.8 \\
RegionGPT~\citeyearpar{guo2024regiongpt}
           & 7B  & 10.6 & 46.4 & 28.5 \\
OMG-LLaVA~\citeyearpar{zhang2024omgllava}
           & 7B  & 5.6 & 32.6 & 19.1 \\
VP-SPHINX~\citeyearpar{lin2025drawunderstand}
           & 13B & 26.3 & 71.6 & 49.0 \\
DAM~\citeyearpar{lian2025describe}
           & 3B  & 52.3 & 82.2 & 67.3 \\
PixelRefer~\citeyearpar{yuan2025pixelrefer}
           & 7B  & 46.8 & 85.4 & 66.1 \\
GAR~\citeyearpar{wang2026gar}
           & 8B  & 50.2 & 84.6 & 67.4 \\
\midrule
\rowcolor{baselinerow}
SAM-Tok~\citeyearpar{zhou2026samtok}
           & 4B  & 46.1 & 85.2 & 65.6 \\
\midrule[0.3pt]
\rowcolor{pixvlrow}
\textbf{PixVL}
           & 4B  & \textbf{53.3} & \textbf{86.6} & \textbf{69.7} \\
\bottomrule
\end{tabularx}

\end{minipage}
\hfill
\begin{minipage}[t]{.485\linewidth}\vspace{0pt}
\caption{Component and reward ablations.}
\label{tab:ablation}
\centering\footnotesize
\setlength{\tabcolsep}{3pt}
\begin{tabularx}{\linewidth}{@{}>{\raggedright\arraybackslash}Xrrr@{}}
\toprule
\rowcolor{sectionblue}
Removed component & GS & Ref & DLC \\
\midrule
Cross-view & 64.4 & 82.5 & 69.0 \\
Cold-start SFT & 63.3 & 82.3 & 67.8 \\
Retrieved confusers & 62.8 & 82.0 & 68.9 \\
Mask-token credit & 64.2 & 82.6 & \textbf{69.7} \\
Quality scaling & 64.4 & 82.3 & 69.5 \\
\rowcolor{pixvlrow}
\textbf{PixVL (full)} & \textbf{64.8} & \textbf{82.7} & \textbf{69.7} \\
\midrule
\rowcolor{sectionlavender}
Reward ablation & Ref & Ref+ & Refg \\
\midrule
Without $r_{\mathrm{choice}}$ & 61.7 & 57.8 & 58.0 \\
\rowcolor{pixvlrow}
\textbf{PixVL (full)} & \textbf{82.7} & \textbf{77.9} & \textbf{78.7} \\
\bottomrule
\end{tabularx}
\par\smallskip
\raggedright\footnotesize
GS: GroundingSuite; Ref/+/g: RefCOCO/+/g; DLC: average accuracy. Each row removes one component under the same model, data, and training budget.
\end{minipage}
\end{table}

\subsection{Ablations and Further Analysis}

\paragraph{Verification.}
Without the semantic reward, RefCOCO/+/g fall to 61.7/57.8/58.0 (Table~\ref{tab:ablation}), demonstrating that geometric rewards alone are insufficient. Among component ablations, removing retrieved confusers causes the largest GroundingSuite drop (64.8 to 62.8), while removing verifier cold-start lowers DLC-Bench from 69.7 to 67.8. Cross-view verification also benefits both metrics. These results support discriminative candidates, an initialized comparison interface, and temporal verification as complementary parts of the cycle. The semantic-reward removal is qualitatively different from the other ablations: it causes a large referring-segmentation collapse, whereas the remaining controls yield incremental gains around a functioning verifier.

\paragraph{Reverse supervision.}
Removing mask-token credit or quality scaling reduces GroundingSuite to 64.2 or 64.4, respectively, supporting both focused credit and reliability-controlled updates. Mask-token credit leaves the reported DLC score at 69.7 while improving segmentation, consistent with directing the reverse loss toward mask prediction. Quality scaling improves all three reported metrics over its ablation, although its effect is smaller than semantic verification. \par

\paragraph{Caption selection and data scale.}
All-caption takes $12\times$ the reported training time, yet changes GroundingSuite/RefCOCO from 64.8/82.7 to 64.5/82.9 and leaves DLC-Bench at 69.7 (Table~\ref{tab:app-caption-selection}). Best-caption therefore avoids repeated reverse-rollout groups without consistent loss of benchmark performance. Separately, only 5k mask-only samples raise GroundingSuite from 56.1 to 61.1 and DLC-Bench from 65.6 to 68.9. Scaling to 250k further improves them to 64.8 and 69.7, with diminishing gains (Appendix~\ref{app:more-experiments}). The improvement beyond initialization supports the use of unpaired masks after extensive paired-data training.

\par
\section{Conclusion}
\label{sec:conclusion}

We introduced PixVL, a self-supervised Mask--Text Consistency Cycle that learns
from mask-only data to jointly improve region understanding and segmentation.
Cross-view reconstruction discourages shortcuts, while hard confusers and
correctness-gated confidence evaluate referring sufficiency beyond pure IoU.
PixVL combines these semantic and geometric rewards for Mask-to-Text learning,
then uses the best caption and its reward to guide Text-to-Mask learning,
assigning segmentation credit only to mask tokens. A shared update turns the
two directions into mutual generators and verifiers, reducing their training trade-off.
PixVL thereby extends pixel-level vision--language alignment beyond scarce
paired mask--text data while improving both capabilities.

\clearpage
\subsection*{AI use statement}
Generative AI tools (ChatGPT/Codex) assisted with adapting this manuscript to the ICLR format, editing prose for clarity, formatting references, and implementing the revised scientific figures. This manuscript-preparation workflow reused the supplied experimental results and did not run new experiments. Experimental results are unchanged. Figure~\ref{fig:iou-failure} is a conceptual illustration; its masks and bars are not measured predictions or scores. The authors are responsible for the final text, claims, and artifacts.

\subsection*{Ethics statement}
\label{sec:ethics}

PixVL inherits the data, privacy, and representation biases of its base MLLM,
segmentation model, and mask-only training sources. Large-scale image and
video masks should therefore be collected and used under appropriate licenses
and privacy safeguards, and the resulting model should be audited across
object categories, demographic attributes, and application domains. Incorrect
captions or masks may provide misleading visual evidence, so high-stakes uses
should retain human verification. The same fine-grained localization ability
may also be misused for unwanted tracking or surveillance; access controls,
dataset governance, and application-specific review are appropriate
mitigations.

\subsection*{Reproducibility statement}
Sections~\ref{sec:problem-formulation} and~\ref{sec:method} specify the model interfaces, verification rewards, rollout construction, advantage coupling, and joint update. Appendix~\ref{app:implementation-details} reports the data mixture, model initialization, training hyperparameters, mask-token format, and prompts. The main experiments and Appendix~\ref{app:more-experiments} provide benchmark results, component ablations, data-scaling analysis, and qualitative comparisons.

\bibliography{references}
\bibliographystyle{iclr2027_conference}
\clearpage
\appendix
\section{Additional Related Work}
\label{app:related-work}

\subsection{Pixel-Level Multimodal Large Language Models}

Visual segmentation has evolved from semantic segmentation, which predicts
category labels for each pixel, to instance and panoptic segmentation, which
requires distinguishing individual objects and their regions. More recently,
referring expression segmentation (RES) extends segmentation from predefined
categories to language-guided object localization, requiring models to identify
the region described by a natural-language query. Along this direction,
open-vocabulary segmentation further connects dense prediction with vision-
language models, enabling segmentation beyond fixed category vocabularies.

The emergence of large vision-language models has further extended segmentation
from a pure perception task to a multimodal interaction capability. Current
pixel-level MLLMs can be broadly categorized according to the information flow
between language and masks: Text-to-Mask models generate pixel regions from
language instructions, Mask-to-Text models interpret user-specified regions,
and unified models support both directions within a single architecture.

For Text-to-Mask segmentation, early works such as LISA introduced reasoning
segmentation by connecting an MLLM with a segmentation decoder through a special
segmentation token. VISA extended this paradigm from images to videos, where
language reasoning must remain consistent across temporal frames. GLaMM
introduced grounded pixel-level understanding by jointly modeling language
generation and dense localization, while InstructSeg unified multiple
instruction-guided segmentation tasks within an MLLM framework. PixelLM explored
pixel-level interaction by explicitly connecting language tokens with mask
prediction. More recent methods investigate reasoning-enhanced and
general-purpose segmentation. Text4Seg reformulated segmentation as language
generation, while LENS optimized segmentation through reinforcement learning
with sentence-, box-, and mask-level rewards. Qwen3-VL-Seg further explores
open-world segmentation ability inherited from large-scale MLLMs.

Although these methods significantly improve language-guided segmentation, they
mainly treat language as the input condition and masks as the final prediction.
Consequently, their training relies heavily on referring expressions,
instruction-mask pairs, or manually designed reasoning annotations, limiting
their scalability when dense language supervision is unavailable.

A complementary direction studies Mask-to-Text region understanding, where a
model receives a visual region and generates descriptions, answers questions,
or performs fine-grained reasoning about the target. RegionGPT and
Draw-and-Understand introduced region-level interaction interfaces for MLLMs,
allowing users to query specific visual regions instead of the entire image.
DAM focused on detailed localized captioning for masked regions in images and
videos. PixelRefer extended region understanding to arbitrary spatio-temporal
granularity, while GAR studied contextual understanding of arbitrary regions.

Compared with image-level captioning, region understanding requires descriptions
that are sufficiently fine-grained to distinguish visually similar instances.
A useful regional description may need category, attributes, spatial relations,
parts, and contextual interactions. Therefore, this direction faces a more
severe limitation of mask-text pair scarcity than conventional image-text
learning.

Recently, several works attempt to unify mask input and mask output in one
architecture. OMG-LLaVA bridges image-, object-, and pixel-level interaction
with a unified representation. Sa2VA combines SAM2 with LLaVA to support dense
grounded understanding across images and videos. UniPixel investigates unified
pixel-level reasoning, while SAM-Tok represents arbitrary masks using compact
mask tokens. Patch-as-Decodable-Token explores decodable visual tokens as a
general interface for dense prediction tasks.

These unified models demonstrate that a single MLLM can technically support
both region understanding and segmentation. However, existing approaches still
optimize the two directions mainly as independent supervised objectives.
Mask-to-Text and Text-to-Mask are trained on separately
collected datasets, with potential interference between their objectives. PixVL differs by treating these two capabilities as mutual
generators and verifiers: mask understanding generates language supervision,
while segmentation provides a verifiable signal for language quality.

\subsection{Self-Training, Cycle Consistency, and Vision-Language
Self-Supervision}

Self-training and pseudo-labeling have long been effective strategies for
learning from unlabeled data. Classical approaches such as Noisy Student
demonstrated that a model can improve itself by generating additional
supervision, provided that pseudo-label quality is controlled. In computer
vision, large-scale segmentation foundation models such as SAM and SAM2 have
made it possible to obtain massive collections of object masks and temporal
mask tracks without dense human annotation. These resources provide abundant
pixel-level supervision but generally lack corresponding natural-language
descriptions.

A straightforward solution is to generate captions for masks and use them as
pseudo mask-text pairs. However, conventional pseudo-labeling assumes that
prediction confidence correlates with label quality, which is difficult for
open-ended regional descriptions. A fluent caption may describe the wrong
instance, omit discriminative attributes, or exploit image-specific shortcuts.

Cycle consistency provides another mechanism for learning without paired
annotations. Previous works such as CycleGAN explored forward-backward
consistency for unpaired translation, while temporal consistency methods use
object persistence across views to learn stable representations. For
pixel-level vision-language learning, the natural cycle is
Mask-to-Text-to-Mask: a model first describes a region and then reconstructs
the corresponding mask from the generated description.

However, unlike traditional cycle-consistency settings, the mapping between
masks and language is inherently many-to-many. A region can have multiple valid
descriptions, while a general description may correspond to multiple regions.
Therefore, geometric reconstruction alone cannot fully measure semantic
correctness. PixVL follows this observation by combining geometric verification
with confuser-based semantic discrimination. Re-segmentation and re-grounding
measure whether the generated description can recover the target region, while
confuser-aware verification evaluates whether the description uniquely identifies
the target among similar candidates.

Cross-view consistency is also closely related to recent representation
learning methods that enforce invariance across augmented observations. Unlike
feature-level contrastive learning, PixVL applies this principle directly to
language generation. By generating descriptions from one temporal view and
verifying them in another view sampled from video mask tracks, PixVL reduces the
risk that the model learns location, scale, or shape shortcuts rather than
semantic region understanding.

Multi-view representation methods also seek to retain useful visual evidence
while reducing redundancy. ZPressor~\citep{wang2025zpressor} compresses
multi-view inputs for scalable feed-forward 3D Gaussian Splatting. PixVL instead
uses temporally separated observations to verify whether regional descriptions
remain discriminative across views.

\subsection{Self-Verification and Reinforcement Learning for Multimodal Models}

Recent advances in large-model post-training increasingly replace expensive
human supervision with automatically verifiable feedback. Self-consistency
methods improve reasoning by aggregating multiple generations, while
self-rewarding approaches use model-based evaluation to construct preference
signals. Reinforcement learning methods such as GRPO further enable optimization
with group-relative rewards when explicit ground-truth labels are unavailable.

In multimodal learning, reinforcement learning has recently been applied to
improve visual reasoning and dense prediction. Vision-R1 explores
vision-language reinforcement learning for reasoning tasks, while LENS introduces
reinforcement learning objectives that jointly optimize language reasoning and
segmentation quality through sentence-, box-, and mask-level rewards.

Nevertheless, regional captioning differs fundamentally from tasks with unique
answers, such as mathematics or classification. A region can have many valid
descriptions, making exact matching unsuitable. Moreover, language quality
depends not only on describing true attributes but also on whether the
description can distinguish the target from visually similar alternatives.

PixVL addresses this challenge by transforming caption evaluation into a
self-verification problem. Given a generated description and a set of hard
confusers, the model evaluates whether the description uniquely identifies the
target region. This provides a semantic verification signal beyond geometric
overlap. Furthermore, PixVL does not optimize only one direction: verified
Mask-to-Text generations are converted into quality-controlled supervision for
Text-to-Mask learning. The selected caption's combined reward controls the strength of reverse
updates, and mask-token-only credit assignment restricts direct segmentation
credit to mask positions. Therefore, PixVL differs from
previous RL-based multimodal post-training by using reinforcement learning not
only to optimize a task objective, but also to establish a self-improving loop
between two complementary pixel-level capabilities.

\section{Implementation Details}
\label{app:implementation-details}

\subsection{Training Configuration}

Table~\ref{tab:app-implementation} summarizes the reported data, rollout,
reward, and optimization configuration. For video samples, the two views are
drawn from the same mask track; image samples reuse the same image and mask.
Confusers are preferentially sampled from nearby valid masks, candidate order
is randomized, and examples without enough valid confusers are discarded.

\begin{table}[!htbp]
\caption{Implementation and training hyperparameters used for PixVL.}
\label{tab:app-implementation}
\centering\footnotesize
\setlength{\tabcolsep}{3pt}
\begin{tabularx}{\linewidth}{@{}lXlX@{}}
\toprule
\multicolumn{2}{l}{\textbf{Data and rollouts}} & \multicolumn{2}{l}{\textbf{Optimization and runtime}} \\
\midrule
MLLM backbone & Qwen3-VL-4B & Optimizer & AdamW \\
Initialization & SAM-Tok-4B & Learning rate & $5\times10^{-6}$ \\
Cold-start data & approximately 20k examples & Weight decay & 0 \\
Cycle-training data & 250000 & Precision & BF16 \\
Image:video ratio & 8:2 & Flash Attention & Enabled \\
Frame interval & 8 frames & Gradient checkpointing & Enabled \\
Confusers & 9 (10-way verification) & Random seed & 20260624 \\
M2T group size & 12 & LoRA rank / alpha & 128 / 256 \\
T2M group size & 12 & LoRA dropout & 0.05 \\
Temperature / top-$p$ & 0.8 / 0.9 & GPUs & 16 \\
Reward weights & $(0.25,0.25,0.5)$ & Per-GPU rollout batch & 8 \\
$(\alpha,\beta,\gamma)$ & & Global batch size & 128 \\
GRPO clipping $\epsilon$ & 0.2 & Gradient accumulation & 1 \\
& & Gradient clipping & global norm 1.0 \\
\midrule
LoRA target modules & \multicolumn{3}{l}{\texttt{q\_proj}, \texttt{k\_proj}, \texttt{v\_proj}, \texttt{o\_proj}, \texttt{gate\_proj}, \texttt{up\_proj}, \texttt{down\_proj}} \\
\bottomrule
\end{tabularx}
\end{table}

Malformed responses, failed mask decoding, and empty masks receive zero
reward, while rollout groups with negligible reward variation are skipped.
The highest-reward caption conditions T2M, whose normalized advantages are
weighted by that caption's verified reward; M2T and T2M gradients are then
combined in one update. The M2T loss covers the full caption, whereas the T2M
loss is restricted to mask-vocabulary positions for mask-token-only credit
assignment.

\subsection{Input and Prompt Formats}

The mask-token input format is
\begin{quote}
\small\ttfamily
<|mt\_start|>\\
<|mt\_xxxx|>\\
<|mt\_yyyy|>\\
<|mt\_end|>
\end{quote}
The caption-generation prompt is
\begin{quote}
\small\ttfamily
Given a detailed description of this region <MASK TOKENS>.\\
Describe it briefly and discriminatively.
\end{quote}
The segmentation prompt is
\begin{quote}
\small\ttfamily
Please segment \{caption\} in this image.
\end{quote}
The same prompt templates are used during rollout generation and policy updates
to avoid a train--inference format mismatch.

\section{Discussion}
\label{app:discussion}

\subsection{From Geometric Reconstructability to Referring Sufficiency}

PixVL begins from the premise that regional caption quality should not be
determined only by whether the original mask can be reconstructed. The more
relevant question is whether the description contains enough information to
identify the target uniquely among similar candidates. Pure IoU compresses all
errors into a final geometric difference and does not identify whether a
low score results from an omitted attribute, a wrong-instance selection, or
an unrelated description. Conversely, an overly broad description can yield
moderate overlap despite lacking target specificity (Figure~\ref{fig:iou-failure}). Choose-one verification explicitly
introduces contrastive alternatives, making the reward respond to how well the
caption separates the target from its confusers.

This does not make IoU unimportant. Re-segmentation IoU and re-grounding box
IoU remain valuable verifiable signals for whether a caption maps to the
correct visual region. PixVL instead reassigns the roles of the three signals:
correctness-gated choose-one confidence measures semantic sufficiency, pixel
IoU measures precise geometric validity, and box IoU provides location
consistency that is more tolerant of boundary noise. Their combination is
better suited to open-ended regional descriptions than any individual metric.

\subsection{From a Training Trade-Off to Bidirectional Cooperation}

Conventional multi-task training treats Mask-to-Text and Text-to-Mask as two
parallel losses over shared parameters. Differences in data volume, gradient
density, and optimization difficulty can create a trade-off. PixVL changes
their relationship through the cycle: Text-to-Mask is not only a task to be
optimized, but also supplies verifiable geometric outcomes for captions;
Mask-to-Text not only generates language, but also supplies its highest-reward
caption as the referring expression for Text-to-Mask. Both objectives are
constructed before their gradients are combined in a single shared update.

This cooperation depends on strict quality control. Independent group normalization compares mask rollouts without retaining a
sample-level measure of caption reliability. Post-normalization caption
weighting controls the strength of this reverse supervision.
If the segmentation reward is assigned to the full response, it also gives
direct credit to language-token predictions unrelated to mask encoding. Cross-view verification,
highest-reward caption selection, reward coupling, and mask-token-only credit
assignment are therefore complementary controls within the learning
loop.

\section{More Experiments}
\label{app:more-experiments}

\subsection{Experiments on Pixel-level Benchmarks}

We further evaluate PixVL on GRES and MR-PACO to assess pixel-level grounding
under generalized referring expressions and multi-round interactions. 
\begin{table}[!htbp]
\begin{minipage}[t]{.47\linewidth}\vspace{0pt}
\caption{GRES validation results. Best reported
scores are bold.}
\label{tab:app-gres}
\centering
\footnotesize
\setlength{\tabcolsep}{1.5pt}
\begin{adjustbox}{max width=\linewidth}
\begin{tabular}{@{}lcccc@{}}
\toprule
\multirow{2}{*}{\textbf{Method}} & \multirow{2}{*}{\textbf{Size}}
& \multicolumn{3}{c}{\textbf{GRES Val}} \\
\cmidrule(lr){3-5}
& & \textbf{gIoU} & \textbf{cIoU} & \textbf{N-acc} \\
\midrule
LISA
    & 7B & 61.6 & 61.8 & 54.7 \\
SAM4MLLM
    & 8B & 71.9 & 67.8 & 66.1 \\
MLLMSeg
    & 8B & \textbf{75.1} & 71.6 & 73.2 \\
HiMTok
    & 8B & 72.1 & 70.4 & -- \\
ARGenSeg
    & 8B & 74.7 & \textbf{72.2} & -- \\
SAMTok
    & 4B & 70.5 & 66.1 & 64.0 \\
\midrule
\rowcolor{pixvlrow}
\textbf{PixVL}
    & 4B & \textbf{75.1} & 70.3 & \textbf{74.1} \\
\bottomrule
\end{tabular}
\end{adjustbox}

\end{minipage}
\hfill
\begin{minipage}[t]{.51\linewidth}\vspace{0pt}
\caption{MR-PACO results (R2--R5: rounds 2--5). Best results are bold.}
\label{tab:app-mr-paco}
\centering
\footnotesize
\setlength{\tabcolsep}{1.5pt}
\begin{adjustbox}{max width=\linewidth}
\begin{tabular}{@{}lcccccc@{}}
\toprule
\multirow{2}{*}{\textbf{Method}} & \multirow{2}{*}{\textbf{Size}}
& \multicolumn{4}{c}{\textbf{MR-PACO}} & \multirow{2}{*}{\textbf{Avg.}} \\
\cmidrule(lr){3-6}
& & \textbf{R2} & \textbf{R3}
& \textbf{R4} & \textbf{R5} & \\
\midrule
LISA
    & 7B & 15.5 & 21.3 & 18.7 & 20.5 & 19.0 \\
SegLMM
    & 7B & 49.7 & 40.9 & 39.4 & 41.9 & 43.0 \\
SAMTok
    & 4B & 53.3 & 44.2 & \textbf{48.2}
    & 47.9 & 48.4 \\
\midrule
\rowcolor{pixvlrow}
\textbf{PixVL} & 4B & \textbf{54.1} & \textbf{45.1} & 47.9 & \textbf{48.7} & \textbf{48.9} \\
\bottomrule
\end{tabular}
\end{adjustbox}

\end{minipage}
\end{table}

\subsection{Ablation Study}
\begin{table}[H]
\caption{Comparison of caption-selection strategies. All-caption uses all $K_c$ M2T rollouts and
therefore incurs $K_c\times$ the training time ($12\times$ in our setting),
while providing only marginal metric changes.}
\label{tab:app-caption-selection}
\centering
\small
\setlength{\tabcolsep}{2pt}
\begin{adjustbox}{max width=\linewidth}
\begin{tabular}{@{}lcccc@{}}
\toprule
\textbf{Method} & \textbf{GroundingSuite} & \textbf{RefCOCO}
& \textbf{DLC--Bench} & \textbf{Train. Time} \\
\midrule
\textbf{Best-caption (Ours)} & \textbf{64.8} & 82.7 & \textbf{69.7} & 1.0$\times$ \\
All-caption & 64.5 & \textbf{82.9} & \textbf{69.7} & $K_c\times$ ($12\times$) \\
\bottomrule
\end{tabular}
\end{adjustbox}

\end{table}

As shown in Table~\ref{tab:app-caption-selection}, using all generated captions
increases the training time by the M2T rollout group size $K_c$ relative to
selecting only the highest-reward caption. With $K_c=12$, All-caption is
$12\times$ as expensive as Best-caption, yet yields only marginal changes on
the three benchmarks. We therefore use Best-caption as the training
implementation throughout the paper.

\begin{figure}[H]
\centering
\includegraphics[width=\linewidth]{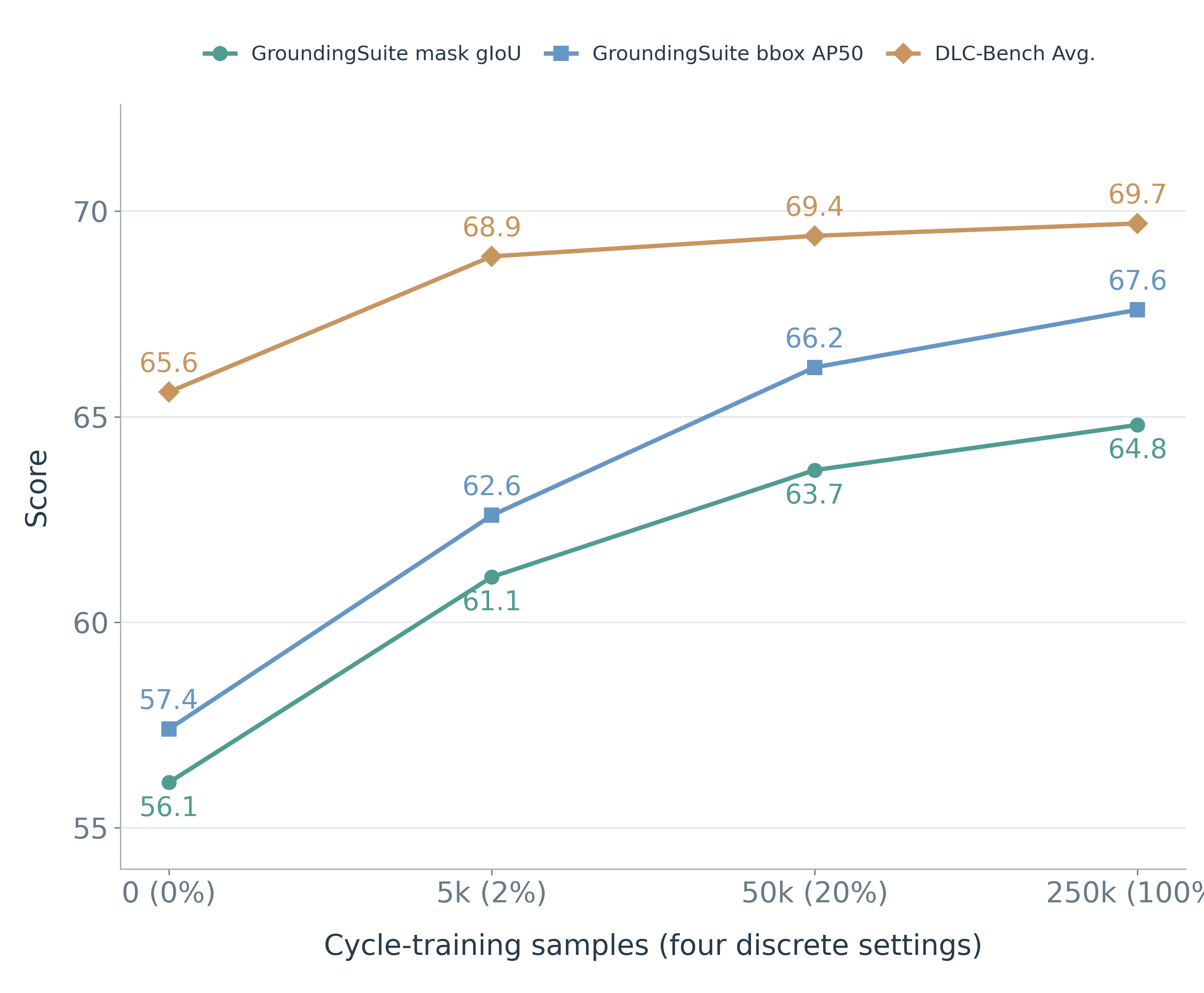}
\caption{Training-data scaling results. Performance improves consistently as
the cycle-training set grows from 0 to 250k examples. The largest gains occur
at smaller data scales, while additional data continues to improve all three
metrics with gradually diminishing returns.}
\label{fig:app-data-scaling}
\end{figure}

Figure~\ref{fig:app-data-scaling} studies the effect of scaling the cycle-
training data from no cycle data to 5k, 50k, and the full 250k examples. All
three metrics improve consistently as more mask-only training data are used.
The largest gain occurs in the first 5k examples: GroundingSuite mask gIoU and
bbox AP50 increase from $56.1$ to $61.1$ and from $57.4$ to $62.6$,
respectively, while the DLC--Bench average rises from $65.6$ to $68.9$.
Scaling from 5k to 250k further improves these metrics to $64.8$, $67.6$, and
$69.7$. The gains become more gradual at larger scales, especially on
DLC--Bench, but remain positive across both region segmentation and region
understanding. This trend supports the scalability of learning bidirectional
pixel--language alignment from mask-only data.

\begin{figure}[H]
\centering
\includegraphics[width=0.98\textwidth,height=0.315\textheight,keepaspectratio]{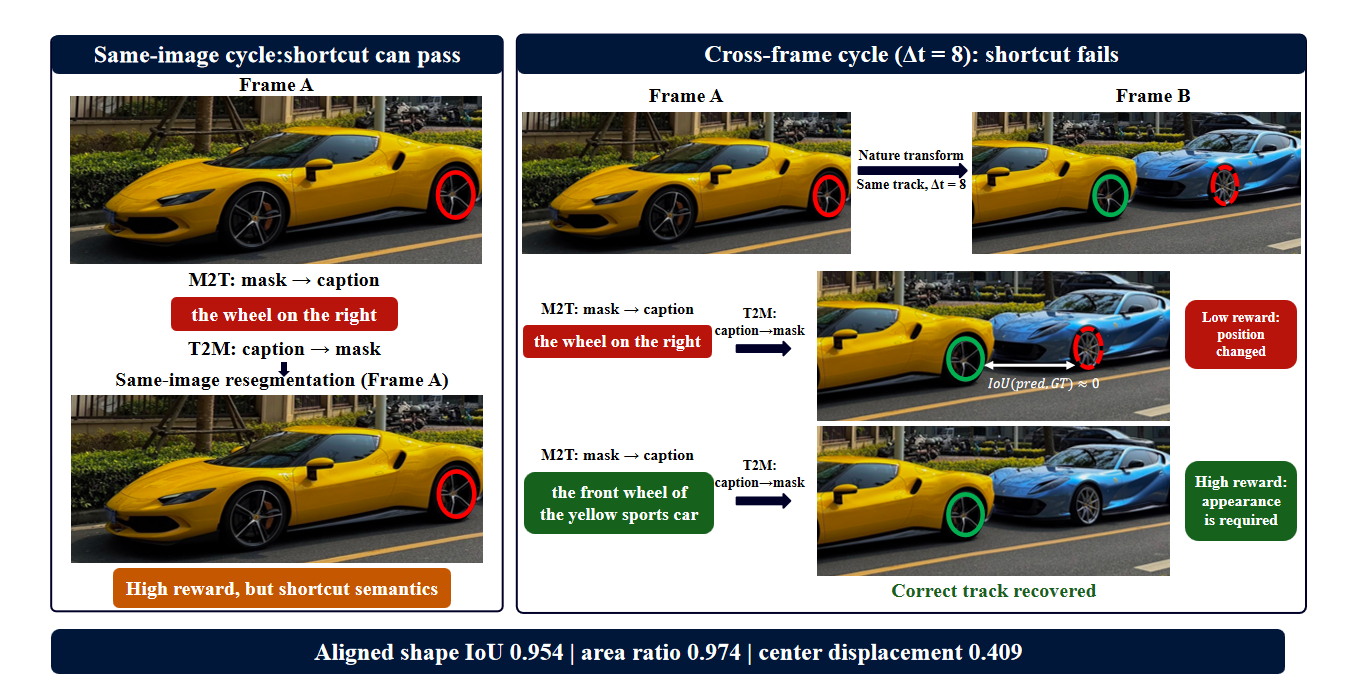}
\caption{Why cross-view verification matters. A same-image cycle can reward
shortcut descriptions based only on position. Across two frames from the same
track, natural scene changes invalidate the positional shortcut, while a
description grounded in the target's appearance still recovers the correct
region.}
\label{fig:app-cross-view}
\end{figure}

Figure~\ref{fig:app-cross-view} illustrates the role of cross-view
verification. In a same-image cycle, a positional description such as ``the
wheel on the right'' can reconstruct the target mask and receive a high reward
without identifying the object's appearance. Natural motion across frames
breaks this shortcut: the same positional phrase points to a different object,
whereas an appearance-grounded description continues to recover the tracked
region. Cross-view verification therefore favors referential semantics that
remain valid under changes in position, scale, and surrounding context.

The two verification paths in the figure highlight where this advantage comes
from. The illustrated views are eight frames apart: although the tracked wheel
retains nearly the same shape and area (aligned shape IoU $0.954$ and area
ratio $0.974$), its center moves substantially and another car enters the
right side of the scene. Consequently, ``the wheel on the right'' switches to
the blue car and produces almost zero overlap with the tracked target. In
contrast, ``the front wheel of the yellow sports car'' recovers the correct
track because it encodes persistent category and appearance evidence. Thus,
cross-view verification acts as a direct stress test for shortcut captions,
rather than merely serving as an additional temporal augmentation.

\subsection{Qualitative Results}

\begin{figure}[H]
\centering
\includegraphics[width=0.98\linewidth,height=0.60\textheight,keepaspectratio]{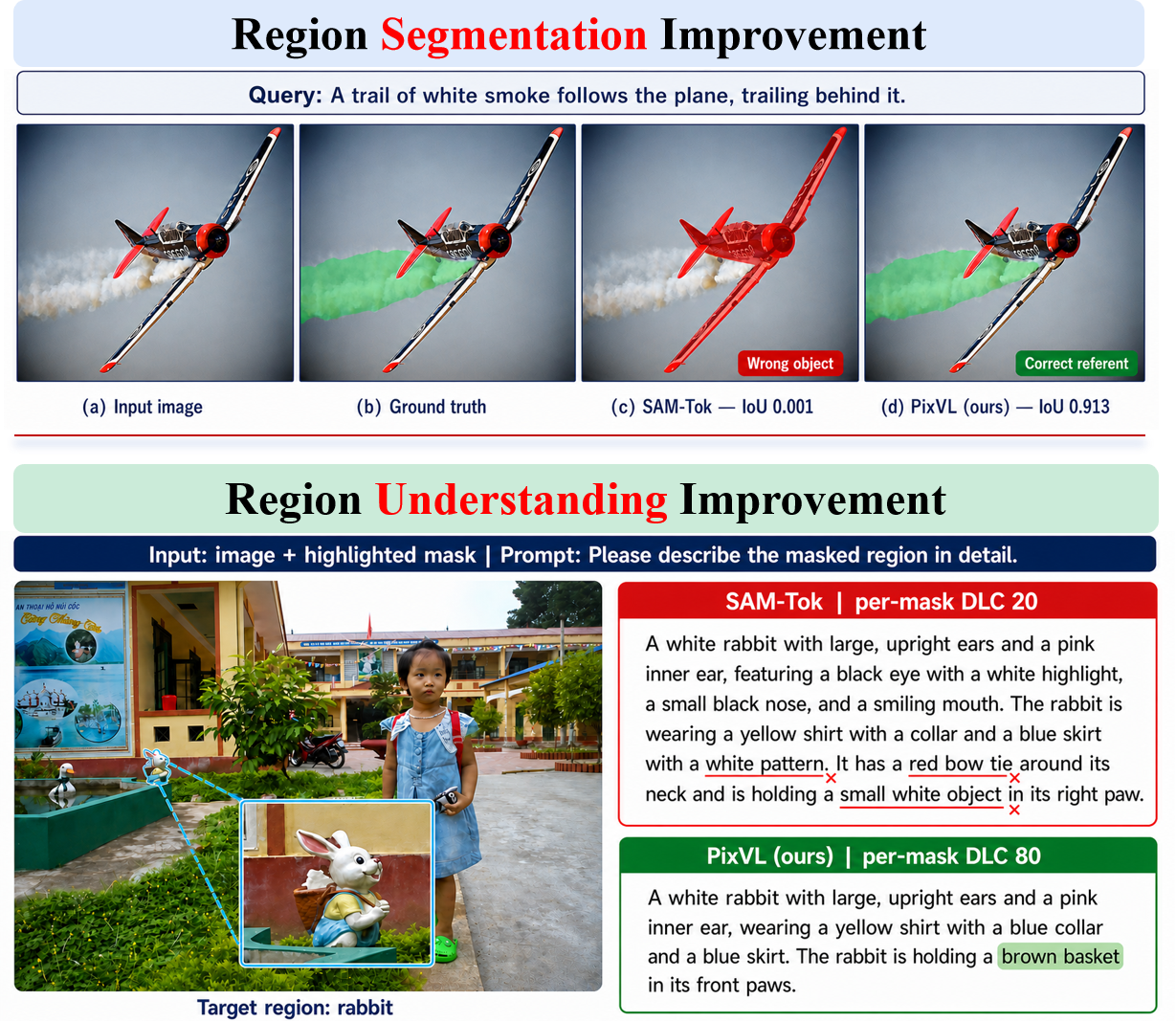}
\caption{Qualitative comparison with SAM-Tok on complementary pixel-level
capabilities. Top: for region segmentation, SAM-Tok selects the salient
airplane, whereas PixVL follows the referring expression and recovers the smoke
trail. Bottom: for region understanding, PixVL gives a more grounded and
target-specific description while avoiding unsupported
attributes.}
\label{fig:app-qualitative}
\end{figure}

Figure~\ref{fig:app-qualitative} compares PixVL with SAM-Tok on the two
complementary pixel-level capabilities. For region segmentation, PixVL follows
the referring expression instead of defaulting to the most salient object. For
region understanding, it produces a more grounded description with fewer
unsupported attributes and more target-specific detail.

\clearpage

In the upper example, visual saliency conflicts with the referring expression.
The airplane is the dominant object, so SAM-Tok produces a plausible but
query-inconsistent airplane mask. The requested region, however, is the less
salient smoke trail. PixVL separates the linguistic target from its salient
source object and traces the elongated smoke region, increasing the per-example
IoU from $0.001$ to $0.913$.

The lower example examines the reverse direction from a selected region to
language. SAM-Tok describes the rabbit but adds several attributes that are not
supported by the target region. PixVL retains the visible, target-specific
evidence, correctly describing the rabbit's clothing and the brown basket in
its front paws without introducing those unsupported details. The per-mask DLC
score consequently increases from $20$ to $80$. Together, the two examples
show that PixVL improves both referential mask prediction and discriminative,
spatially grounded region description.

\begin{figure}[H]
\centering
\includegraphics[width=\linewidth]{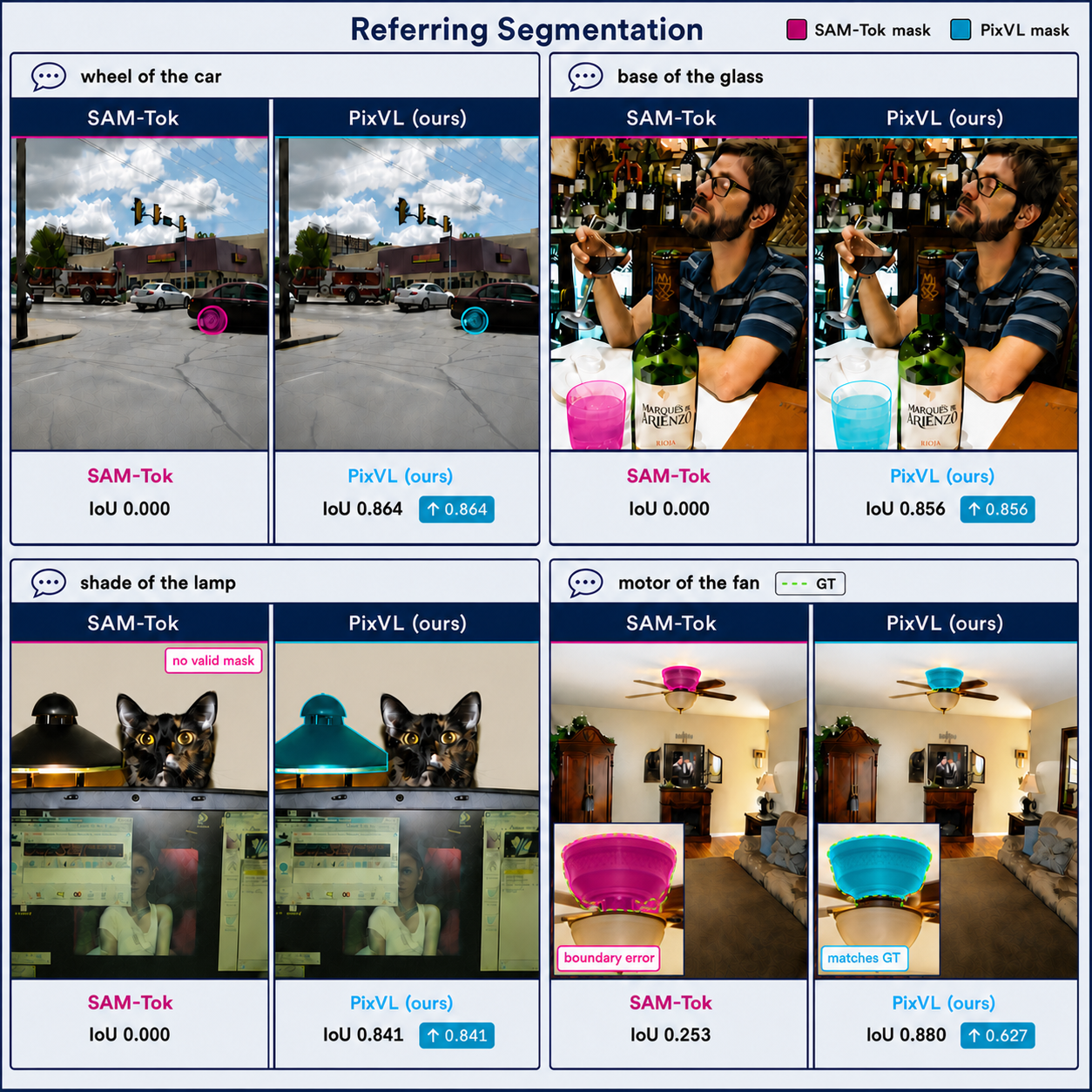}
\caption{More qualitative results for region segmentation. Across four
fine-grained referring expressions, PixVL follows the requested region more
accurately than SAM-Tok, including small object parts and cases in which
SAM-Tok produces no valid mask or inaccurate boundaries.}
\label{fig:app-more-segmentation}

\end{figure}

\begin{figure}[H]
\centering
\includegraphics[width=\linewidth]{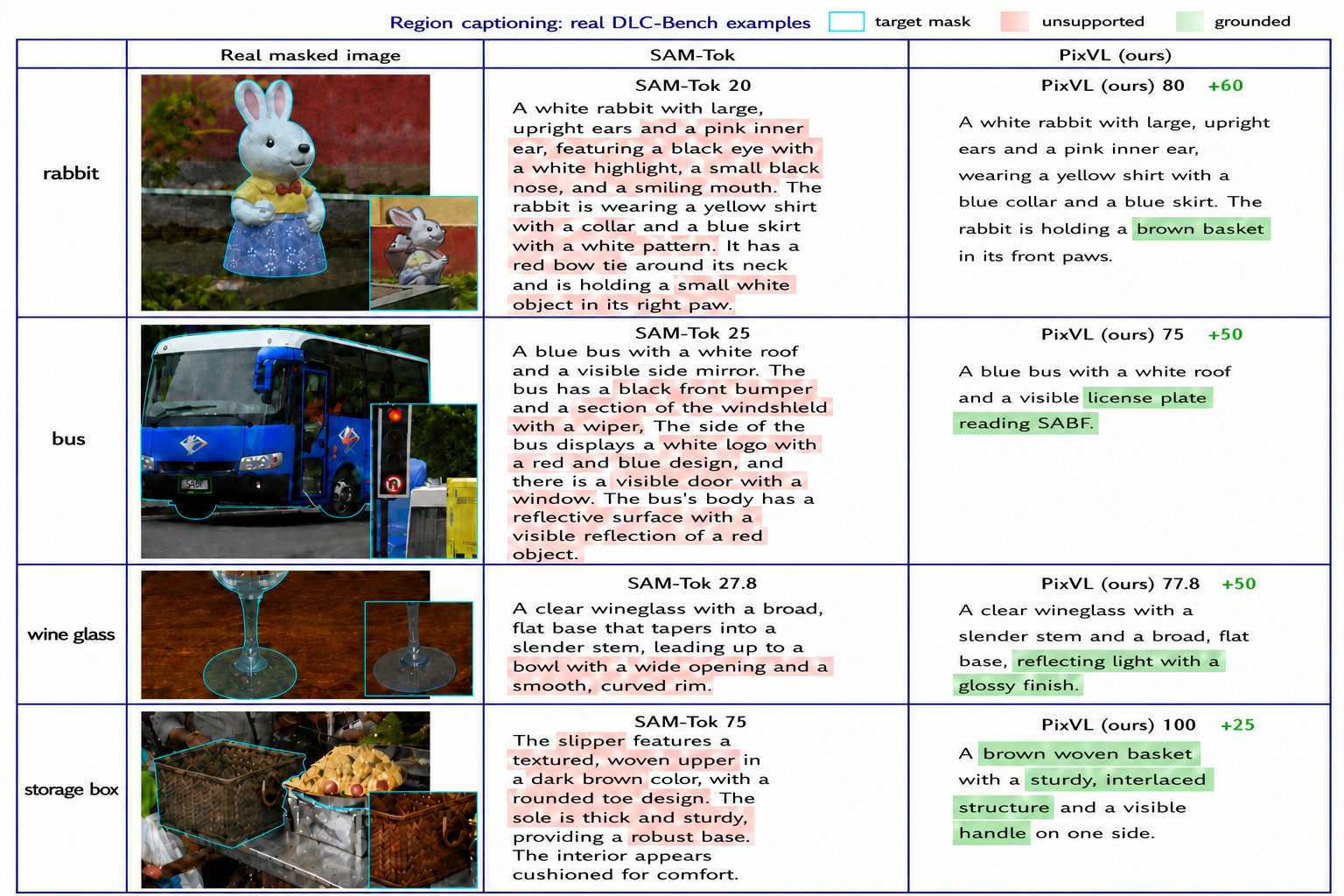}
\caption{More qualitative results for region understanding on real
DLC--Bench examples. Compared with SAM-Tok, PixVL produces more grounded,
target-specific descriptions with fewer unsupported attributes. Green and red
highlights denote grounded and unsupported content, respectively.}
\label{fig:app-more-understanding}
\end{figure}

Figures~\ref{fig:app-more-segmentation} and
\ref{fig:app-more-understanding} provide additional qualitative comparisons
for region segmentation and region understanding, respectively.

\clearpage

The examples in Figure~\ref{fig:app-more-segmentation} concern a car wheel, the base of a glass, a lamp shade, and a fan motor. These queries require separating a specific part from its surrounding object. The comparison includes both a missing prediction and masks that select an incorrect region or boundary; PixVL recovers the requested part in each illustrated case. These are per-example qualitative comparisons, complementing the aggregate benchmark results above.

\clearpage

\section{Limitations}
\label{app:limitations}

Although PixVL does not require human-written mask--text pairs during its
large-scale cycle stage, it still depends on image--mask annotations, video
mask tracks, or masks generated by an automatic segmentation model. It is
therefore more precise to describe the approach as self-supervised language
alignment on mask-only data rather than fully unsupervised learning. The
choose-one cold start also introduces limited initialization supervision. Its
role is to establish a reliable verification protocol rather than directly
cover the large space of regional descriptions.

The method has several further limitations. Confuser coverage bounds the reward:
a caption that distinguishes the current candidate set is not guaranteed to
remain sufficient in an open-world setting. Video tracks may contain
occlusion, identity switches, or mask noise and therefore require robust pair
filtering. Each example needs multiple caption rollouts,
re-segmentation checks, choose-one inference, and potentially multiple T2M
rollouts, making training more expensive than ordinary SFT. Shared visual
features, an offline verifier, distillation, or hierarchical filtering could
reduce this cost.

Finally, cross-view pairs weaken position and shape shortcuts but cannot
guarantee their complete elimination. More systematic analysis of coordinate
terms, caption attribute types, cross-domain transfer, and robustness to
automatically generated masks is needed to confirm that the learned regional
semantics generalize.

\end{document}